\documentclass{article}
\usepackage[utf8]{inputenc}
\usepackage{amsmath,amssymb}
\usepackage{graphicx}
\usepackage{booktabs}
\usepackage{hyperref}
\usepackage{caption}
\usepackage{subcaption}
\usepackage[margin=1in]{geometry}
\usepackage{float}  
\usepackage{authblk}  

\title{Evaluating Large Language Models on Quantum Mechanics: \\ A Comparative Study Across Diverse Models and Tasks}

\author[1,2]{S. K. Rithvik\thanks{Corresponding author. Email: rithvik\_ks@iitgn.ac.in}}
\affil[1]{Quantum Science and Technology Laboratory, Physical Research Laboratory, Navrangpura, Ahmedabad 380009, India}
\affil[2]{Indian Institute of Technology Gandhinagar, Palaj, Gandhinagar 382055, India}

\date{\today}

\begin{document}

\maketitle

\begin{abstract}
We present a systematic evaluation of large language models on quantum mechanics problem-solving. Our study evaluates 15 models from five providers (OpenAI, Anthropic, Google, Alibaba, DeepSeek) spanning three capability tiers on 20 tasks covering derivations, creative problems, non-standard concepts, and numerical computation, comprising 900 baseline and 75 tool-augmented assessments. Results reveal clear tier stratification: flagship models achieve 81\% average accuracy, outperforming mid-tier (77\%) and fast models (67\%) by 4pp and 14pp respectively. Task difficulty patterns emerge distinctly: derivations show highest performance (92\% average, 100\% for flagship models), while numerical computation remains most challenging (42\%). Tool augmentation on numerical tasks yields task-dependent effects: modest overall improvement (+4.4pp) at 3x token cost masks dramatic heterogeneity ranging from +29pp gains to -16pp degradation. Reproducibility analysis across three runs quantifies 6.3pp average variance, with flagship models demonstrating exceptional stability (GPT-5 achieves zero variance) while specialized models require multi-run evaluation. This work contributes: (i) a benchmark for quantum mechanics with automatic verification, (ii) systematic evaluation quantifying tier-based performance hierarchies, (iii) empirical analysis of tool augmentation trade-offs, and (iv) reproducibility characterization. All tasks, verifiers, and results are publicly released.
\end{abstract}

\section{Introduction}

Large language models (LLMs) have demonstrated remarkable capabilities across diverse domains, from natural language understanding~\cite{brown2020language} to mathematical reasoning~\cite{cobbe2021training} and code generation~\cite{chen2021evaluating}. However, their ability to handle advanced scientific concepts, particularly in quantum mechanics—a field requiring both symbolic manipulation and numerical computation—remains underexplored. As LLMs increasingly serve as research assistants and educational tools in scientific contexts, systematic evaluation of their capabilities and limitations in specialized domains becomes essential.

Quantum mechanics presents distinct evaluation challenges compared to existing benchmarks. While prior work has assessed LLMs on mathematical reasoning (GSM8K~\cite{cobbe2021training}, MATH~\cite{hendrycks2021math}), coding (HumanEval~\cite{chen2021evaluating}, MBPP~\cite{austin2021mbpp}), and broad scientific knowledge (MMLU~\cite{hendrycks2021measuring}), quantum mechanics demands integration of multiple cognitive modes: understanding concepts that defy classical intuition, executing multi-step symbolic derivations with operator algebra, applying design principles under physical constraints, and implementing numerical algorithms for computational predictions. Unlike domain-general benchmarks, quantum mechanics evaluation requires simultaneous assessment of conceptual knowledge, mathematical precision, and computational thinking.

Recent advances in tool augmentation~\cite{schick2023toolformer} have enabled LLMs to leverage external capabilities like code execution, yet their impact on scientific problem-solving remains incompletely characterized. While tool augmentation shows promise for mathematical tasks, scientific domains present unique requirements—numerical stability considerations, algorithm selection, multi-step workflows—that may benefit differently from computational tools compared to analytical reasoning. Understanding when and how tool augmentation enhances (or degrades) performance informs effective deployment strategies for AI-assisted scientific computing.

This work provides the first comprehensive evaluation of LLMs on quantum mechanics tasks, addressing three research questions: \textbf{RQ1:} How do state-of-the-art LLMs perform across diverse quantum mechanics task categories? \textbf{RQ2:} Does tool augmentation with code execution improve performance on numerical quantum problems, and at what cost? \textbf{RQ3:} How reproducible are LLM responses on quantum mechanics tasks across multiple runs?

We evaluate 15 LLMs spanning three performance tiers (fast, mid-tier, flagship) on 20 tasks across four categories: derivations (symbolic operator manipulation), creative design (optimization under physical constraints), non-standard concepts (phenomena understanding), and numerical computation (algorithmic implementation or physics-based estimation). Our evaluation protocol includes 900 baseline assessments with three independent runs per model-task combination, plus 75 tool-augmented evaluations enabling Python code execution for numerical tasks.

Our contributions include: (1) a quantum mechanics benchmark with 20 tasks and automatic verification covering diverse cognitive requirements, (2) systematic evaluation of 15 models across three performance tiers with comprehensive cost, token usage, and inference time metrics, (3) tool augmentation analysis quantifying task-specific effects of code execution on accuracy and resource consumption, (4) reproducibility quantification across three independent runs at temperature zero, and (5) public release of all tasks, verifiers, and evaluation infrastructure to enable community extensions.

\section{Methods}

Our evaluation methodology comprises three components: model selection across capability tiers and providers, task design spanning derivations to numerical computation, and evaluation protocols for both baseline and tool-augmented assessments. This section details each component and the rationale behind our design choices.

\subsection{Model Selection}

The rapid evolution of large language models presents both opportunities and challenges for scientific benchmarking. With new models released monthly and significant architectural diversity across providers, selecting a representative evaluation set requires balancing comprehensiveness with practical constraints. Our approach prioritizes four criteria: (1) \textbf{provider diversity} to capture different training philosophies and data curation strategies, (2) \textbf{architectural variety} spanning parameter scales from 32B to 671B and including both standard and reasoning-augmented architectures, (3) \textbf{performance stratification} across capability tiers to understand cost-accuracy trade-offs, and (4) \textbf{temporal relevance} focusing on models available in early 2025 to reflect current state-of-the-art.

We evaluated 15 LLMs organized into three performance tiers reflecting their position in the capability-cost landscape:

\textbf{Fast tier:} Claude 3.5 Haiku~\cite{anthropic2024claude35}, GPT-3.5 Turbo~\cite{openai2023gpt35}, Gemini 2.0 Flash~\cite{google2024gemini2}, Qwen 2.5 Coder 32B~\cite{qwen2024qwen25coder}, DeepSeek R1 Distill 32B~\cite{deepseek2025r1}. These models prioritize inference speed and cost efficiency, making them suitable for high-throughput applications where computational budget constraints are primary. Models in this tier have approximately 30B parameters (confirmed for Qwen 2.5 Coder 32B and DeepSeek R1 Distill 32B; Claude 3.5 Haiku and GPT-3.5 Turbo parameter counts are undisclosed but widely believed to be in this range based on inference characteristics).

\textbf{Mid-tier:} Claude Sonnet 4~\cite{anthropic2024claude4}, GPT-4o~\cite{openai2024gpt4o}, Gemini 2.5 Flash~\cite{google2025gemini25}, Qwen3 235B~\cite{qwen2025qwen3}, DeepSeek V3~\cite{deepseek2024v3}. This tier represents the current workhorses of AI-assisted scientific computing, balancing strong performance with manageable costs for sustained research use. Models in this tier cluster around 200B parameters (confirmed for Qwen3 235B and DeepSeek V3 671B; Claude Sonnet 4 and GPT-4o parameter counts are undisclosed but widely believed to be approximately 200B based on capability profiles and deployment patterns). DeepSeek V3 was selected as the mid-tier representative from DeepSeek to maintain one model per provider per tier, with DeepSeek R1 positioned as their flagship offering. These models are widely deployed in academic and industrial research environments.

\textbf{Flagship tier:} Claude Sonnet 4.5~\cite{anthropic2024claude45}, GPT-5~\cite{openai2024gpt5}, Gemini 2.5 Pro~\cite{google2025gemini25pro}, Qwen3 Max~\cite{qwen2025qwen3max}, DeepSeek R1~\cite{deepseek2025r1}. These cutting-edge models represent the frontier of AI capabilities as of early 2025. This tier includes recently released flagship models (GPT-5 released December 2024, Qwen3 Max released January 2025) and architectures with extended reasoning capabilities (DeepSeek R1 with reinforcement learning from reasoning traces).

This 15-model selection provides systematic $5 \times 3$ coverage with exactly one model per provider per tier (Anthropic: 3.5 Haiku/Sonnet 4/Sonnet 4.5; OpenAI: GPT-3.5/GPT-4o/GPT-5; Google: Flash 2.0/Flash 2.5/Pro 2.5; Alibaba: Qwen2.5 Coder 32B/Qwen3 235B/Qwen3 Max; DeepSeek: R1 Distill/V3/R1). This balanced design enables direct provider comparisons across tiers while capturing intra-tier performance variance from architectural diversity, training methodologies, and specialization strategies. The selection includes both closed-source API-only models (GPT-5, Claude Sonnet 4.5) and open-weights models (Qwen3 235B, DeepSeek V3/R1) to reflect the full spectrum of deployment scenarios in scientific computing.

\subsection{Task Design}

We designed 20 quantum mechanics tasks across four categories, each testing distinct cognitive capabilities. Our benchmark emphasizes task quality and cognitive diversity: each task undergoes expert validation with verified ground truth and automatic verification systems, enabling rigorous assessment across symbolic reasoning, design synthesis, conceptual knowledge, and computational capabilities. With 900 baseline evaluations (15 models × 20 tasks × 3 runs) plus 75 tool-augmented assessments, our evaluation provides sufficient statistical power to establish tier-based performance hierarchies (Section~\ref{sec:results}) and quantify reproducibility bounds. All tasks employ \textbf{consistent multiple-choice format} (A/B/C/D) enabling objective evaluation across diverse LLM architectures while preserving assessment of reasoning through free-form explanations preceding answer selection.

\textbf{Task categories differ in conceptual focus and problem structure:}
\begin{itemize}
\item \textbf{Derivations (D):} Symbolic reasoning and operator algebra requiring derivation of correct mathematical expressions from commutators, uncertainty relations, unitary transformations, perturbation theory, and entropy maximization
\item \textbf{Creative (C):} Design optimization and theoretical limits testing understanding of POVM discrimination, entanglement witnesses, error correction, circuit design, and quantum advantages  
\item \textbf{Non-standard (N):} Conceptual understanding of non-standard quantum phenomena including PT-symmetric systems, quantum thermodynamics, resource theories, topological computing, and quantum metrology
\item \textbf{Numerical (T)}: Computational problems with code execution support testing eigenstate decomposition, quantum tunneling, entanglement evolution, variational methods, and open system dynamics. These tasks uniquely enable dual evaluation modes: with computational tools (testing numerical implementation capabilities) and without tools (testing physics intuition and order-of-magnitude reasoning for quantitative predictions).
\end{itemize}

This design distinguishes \textbf{symbolic reasoning} (D tasks: operator manipulation), \textbf{design synthesis} (C tasks: optimization under constraints), \textbf{conceptual knowledge} (N tasks: phenomena understanding), and \textbf{computational capability} (T tasks: numerical workflows or physics-based estimation). The T tasks specifically test both computational implementation skills and physics intuition, enabling assessment of whether models can arrive at correct numerical answers through analytical reasoning when computational tools are unavailable. All categories employ multiple-choice verification ensuring reproducible assessment.

\subsubsection{Derivation Tasks (D1--D5)}

Derivation tasks assess symbolic reasoning and algebraic manipulation through multiple-choice questions requiring derivation of correct expressions~\cite{sakurai2017quantum,griffiths2018quantum}. These tasks evaluate whether models can: (1) apply commutator identities and Pauli algebra, (2) manipulate quantum operators symbolically, (3) recognize correct mathematical forms, and (4) distinguish between algebraically similar but distinct expressions.

\textbf{D1 (Commutator Algebra):} Compute the commutator $[\sigma_z, B(\theta)]$ where $B(\theta) = \cos(\theta)\sigma_x + \sin(\theta)\sigma_y$ is a rotated spin operator. Models must derive the symbolic form using Pauli commutation relations $[\sigma_z, \sigma_x] = 2i\sigma_y$ and $[\sigma_z, \sigma_y] = -2i\sigma_x$.

\textit{Options:} \textbf{(A)} $2[\cos(\theta)\sigma_x + \sin(\theta)\sigma_y]$; \textbf{(B)} $2i[\cos(\theta)\sigma_y - \sin(\theta)\sigma_x]$; \textbf{(C)} Zero; \textbf{(D)} $2i[\sin(\theta)\sigma_x + \cos(\theta)\sigma_y]$.

\textit{Correct answer:} \textbf{B} ($2i[\cos(\theta)\sigma_y - \sin(\theta)\sigma_x]$). Using linearity: $[\sigma_z, B(\theta)] = \cos(\theta)[\sigma_z, \sigma_x] + \sin(\theta)[\sigma_z, \sigma_y] = 2i[\cos(\theta)\sigma_y - \sin(\theta)\sigma_x]$. Option A misses imaginary unit, C incorrectly assumes commutation, D has wrong signs/terms. Tests Pauli algebra and SU(2) structure.

\textbf{D2 (Uncertainty Relations):} For state $|\psi\rangle = \cos(\alpha/2)|\uparrow\rangle + \sin(\alpha/2)|\downarrow\rangle$ where $|\uparrow\rangle, |\downarrow\rangle$ are $\sigma_z$ eigenstates, determine uncertainty $\Delta\sigma_y = \sqrt{\langle\sigma_y^2\rangle - \langle\sigma_y\rangle^2}$. Models must compute expectation values using Pauli properties.

\textit{Options:} \textbf{(A)} $|\sin(\alpha)|$; \textbf{(B)} $|\cos(\alpha)|$; \textbf{(C)} $\sqrt{1 - \cos^2(\alpha)}$; \textbf{(D)} $1$.

\textit{Correct answer:} \textbf{D} ($\Delta\sigma_y = 1$, constant for all $\alpha$). Key insight: $\langle\sigma_y\rangle = 0$ for all $\alpha$ (state in $xz$-plane) and $\langle\sigma_y^2\rangle = 1$ (Pauli identity $\sigma_y^2 = I$), yielding maximum uncertainty in $y$-direction. Options A/B/C incorrectly suggest $\alpha$-dependence. Tests Pauli identities and geometric understanding of quantum states.

\textbf{D3 (Unitary Transformations):} Find unitary operator $U$ such that $U^\dagger\sigma_z U = \sigma_x$ (basis transformation from computational to Hadamard basis). Models must identify correct matrix achieving this transformation.

\textit{Options:} \textbf{(A)} $U = \frac{1}{\sqrt{2}}\begin{bmatrix}1 & 1 \\ 1 & -1\end{bmatrix}$; \textbf{(B)} $U = \begin{bmatrix}1 & 0 \\ 0 & i\end{bmatrix}$; \textbf{(C)} $U = \begin{bmatrix}0 & 1 \\ 1 & 0\end{bmatrix}$; \textbf{(D)} $U = \frac{1}{\sqrt{2}}\begin{bmatrix}1 & -i \\ -i & 1\end{bmatrix}$.

\textit{Correct answer:} \textbf{A} (Hadamard gate $H = \frac{1}{\sqrt{2}}\begin{bmatrix}1 & 1 \\ 1 & -1\end{bmatrix}$). This unitary transforms $Z$-eigenstates to $X$-eigenstates through $\pi/2$ rotation about $(X+Z)/\sqrt{2}$ axis. Option B is phase gate (no rotation), C gives $U^\dagger\sigma_z U = -\sigma_z$, D has improper structure. Tests basis transformations and operator conjugation.

\textbf{D4 (Perturbation Theory):} For Hamiltonian $H = \sigma_z + \lambda\sigma_x$ with $\lambda=0.1$, compare exact eigenvalues (from diagonalization $E_\pm = \pm\sqrt{1 + \lambda^2}$) with second-order perturbation theory predictions $E_\pm \approx \pm(1 + \lambda^2/2)$. Models must evaluate numerical accuracy of approximation.

\textit{Options:} \textbf{(A)} $\sim 1\%$; \textbf{(B)} $\sim 5\%$; \textbf{(C)} $< 0.01\%$; \textbf{(D)} $> 10\%$.

\textit{Correct answer:} \textbf{C} (relative error $< 0.01\%$). Exact: $E_+ = \sqrt{1.01} \approx 1.00499$. Perturbative: $E_+^{(2)} = 1.005$. Relative error $\approx |1.005 - 1.00499|/1.00499 \approx 0.001\%$, demonstrating excellent second-order accuracy at $\lambda \ll 1$. Options A/B/D overestimate error (would apply at larger $\lambda$). Tests understanding of perturbation theory convergence.

\textbf{D5 (Entropy Maximization):} For qubit density matrix $\rho = p|0\rangle\langle 0| + (1-p)|1\rangle\langle 1|$ with von Neumann entropy $S(\rho) = -\text{Tr}(\rho \log \rho) = -p\log(p) - (1-p)\log(1-p)$, find value of $p \in [0,1]$ maximizing entropy. Models must apply optimization to quantum information.

\textit{Options:} \textbf{(A)} $p = 0$; \textbf{(B)} $p = 1/2$; \textbf{(C)} $p = 1$; \textbf{(D)} $p = 1/\sqrt{2}$.

\textit{Correct answer:} \textbf{B} ($p = 1/2$). Taking derivative: $\frac{dS}{dp} = \log\frac{1-p}{p} = 0 \Rightarrow p = 1/2$, yielding $S_{\max} = \log 2 = 1$ bit (maximally mixed state). Options A/C are pure states with $S=0$, D confuses probability with amplitude. Tests quantum information fundamentals and recognition that maximum uncertainty corresponds to equal mixture.

\textbf{Verification Method:} All models receive a system prompt requiring responses to end with "FINAL ANSWER: [letter]" format. Automated verifiers extract answers with priority order: (1) "FINAL ANSWER: X" pattern, (2) JSON \texttt{\{"answer": "X"\}}, (3) fallback patterns like "Answer: X", (4) last standalone A/B/C/D. Binary scoring: 1 point for correct match, 0 otherwise. Models can show free-form reasoning before the final answer declaration.

\textbf{Category Characteristics:} Derivations test symbolic reasoning, operator algebra, and quantum mechanical principles through focused questions requiring identification of correct mathematical expressions. Multiple-choice format ensures objective evaluation while distractors probe common conceptual errors and calculation mistakes.

\subsubsection{Creative Tasks (C1--C5)}

Creative tasks evaluate design and construction capabilities through multiple-choice questions about optimal solutions, testing understanding of quantum measurement theory, entanglement, error correction, and quantum advantages~\cite{nielsen2010quantum,preskill2015quantum}.

\textbf{C1 (POVM Design):} Design 3-outcome positive operator-valued measure (POVM) optimizing state discrimination between $|0\rangle$ and $|+\rangle = (|0\rangle+|1\rangle)/\sqrt{2}$. POVM elements $\{E_1, E_2, E_3\}$ must satisfy positive semidefiniteness and completeness $E_1 + E_2 + E_3 = I$. Models must determine maximum achievable discrimination measure $d = p(1|0\rangle) - p(1|+\rangle)$ where $p(i|\psi) = \langle\psi|E_i|\psi\rangle$.

\textit{Options:} \textbf{(A)} $d = 0.5$; \textbf{(B)} $d = 0.15$; \textbf{(C)} $d = 0.4$; \textbf{(D)} $d = 0.25$.

\textit{Correct answer:} \textbf{A} ($d = 0.5$). Optimal POVM uses projectors $E_1 = |0\rangle\langle 0|$, $E_2 = |1\rangle\langle 1|$, $E_3 = 0$, yielding $p(1|0\rangle) = 1.0$ and $p(1|+\rangle) = 0.5$, thus $d = 0.5$. This is maximum achievable for these non-orthogonal states. Distractors represent suboptimal POVM designs that fail to maximize discrimination. Tests quantum measurement optimization and fundamental limits of state discrimination.

\textbf{C2 (Entanglement Witness):} Construct Hermitian operator $W$ detecting Bell state $|\Phi^+\rangle = (|00\rangle+|11\rangle)/\sqrt{2}$ while remaining non-negative on separable states. Witness must satisfy $\text{Tr}(W\rho) \geq 0$ for all separable states $\rho$ but $\text{Tr}(W|\Phi^+\rangle\langle\Phi^+|) < 0$. Models must determine expectation value on Bell state for ideal witness.

\textit{Options:} \textbf{(A)} $-0.5$; \textbf{(B)} $0$; \textbf{(C)} $-1$; \textbf{(D)} $-0.25$.

\textit{Correct answer:} \textbf{A} ($\text{Tr}(W|\Phi^+\rangle\langle\Phi^+|) = -0.5$). Standard witness construction $W = I/2 - |\Phi^+\rangle\langle\Phi^+|$ yields $\text{Tr}(W|\Phi^+\rangle\langle\Phi^+|) = 1/2 - 1 = -0.5$ while maintaining $\text{Tr}(W|00\rangle\langle 00|) = 0 \geq 0$ for separable states. Option B would fail to detect entanglement, C would violate normalization, D is suboptimal. Tests entanglement theory and operator design.

\textbf{C3 (Error Correction):} Design quantum code protecting against single bit-flip errors on 3-qubit system. Code encodes logical states $|0\rangle_L$ and $|1\rangle_L$ using 3 physical qubits and must detect and correct errors $X_1$, $X_2$, or $X_3$ through syndrome measurements. Models must determine how many distinct syndrome outcomes are needed to distinguish between no error and three possible single-qubit errors.

\textit{Options:} \textbf{(A)} 2; \textbf{(B)} 3; \textbf{(C)} 4; \textbf{(D)} 8.

\textit{Correct answer:} \textbf{C} (4 distinct outcomes). Must distinguish 4 cases: $\{\text{no error}, X_1, X_2, X_3\}$, requiring 4 distinct syndromes. Standard 3-qubit code uses stabilizers $Z_1Z_2$ and $Z_2Z_3$ giving syndrome pairs: $(0,0) \to$ no error, $(1,0) \to X_1$, $(1,1) \to X_2$, $(0,1) \to X_3$. Option A cannot distinguish 4 cases, B conflates number of errors with syndromes, D would be full tomography. Tests quantum error correction fundamentals and syndrome measurement theory.

\textbf{C4 (Variational Ansatz):} Design parameterized quantum circuit achieving full Bloch sphere coverage (any single-qubit pure state). Starting from $|0\rangle$, circuit must use rotation gates with adjustable parameters to reach arbitrary state $|\psi\rangle = \cos(\theta/2)|0\rangle + e^{i\phi}\sin(\theta/2)|1\rangle$. Models must determine minimum number of parameters required.

\textit{Options:} \textbf{(A)} 2; \textbf{(B)} 3; \textbf{(C)} 4; \textbf{(D)} 5.

\textit{Correct answer:} \textbf{A} or \textbf{B} (both accepted). Two parameters $(\theta, \phi)$ are mathematically sufficient using circuit $R_z(\phi)R_y(\theta)|0\rangle$ to reach any pure state. Three-parameter Euler decomposition $R_y(\alpha)R_x(\beta)R_y(\gamma)$ is standard convention for arbitrary single-qubit unitaries. Task is ambiguous: strict minimum is 2, but conventional answer is 3. Tests quantum circuit design and Bloch sphere parameterization understanding.

\textbf{C5 (CHSH Quantum Advantage):} Design quantum protocol for CHSH nonlocal game. Alice receives input $x \in \{0,1\}$, Bob receives $y \in \{0,1\}$, they output bits $a, b \in \{0,1\}$ with win condition $a \oplus b = x \cdot y$. Using shared entanglement and local measurements, models must determine maximum quantum winning probability.

\textit{Options:} \textbf{(A)} 0.85; \textbf{(B)} 0.75; \textbf{(C)} 0.50; \textbf{(D)} 0.875.

\textit{Correct answer:} \textbf{A} (maximum quantum probability $\approx 0.85$). Optimal strategy uses Bell state with rotated measurements yielding $P_{\text{win}} = \cos^2(\pi/8) = (2+\sqrt{2})/4 \approx 0.8536$, corresponding to CHSH parameter $S = 2\sqrt{2}$ (Tsirelson bound). Option B is classical bound (no advantage), C is random guessing, D exceeds quantum mechanics limit. Most challenging creative task, testing understanding of quantum nonlocality and Bell inequalities.

\textbf{Verification Method:} Identical to derivation tasks—system prompt enforces "FINAL ANSWER: [letter]" format with multi-pattern extraction and binary scoring. Models provide reasoning before final answer.

\textbf{Category Characteristics:} Creative tasks test design optimization, constraint satisfaction, and understanding of quantum advantages through focused questions about optimal parameters and theoretical limits. Multiple-choice format ensures reproducible evaluation of creative problem-solving across diverse LLM architectures.

\subsubsection{Non-Standard Concepts (N1--N5)}

Non-standard tasks test knowledge of advanced topics from contemporary quantum research that are typically absent from standard graduate textbooks. These are multiple-choice questions with carefully crafted distractors targeting common misconceptions.

\textbf{N1 (PT-Symmetric Hamiltonians):} Analyze non-Hermitian Hamiltonian $H = p^2/(2m) + iV_0x$ with formal PT-symmetry. Question asks which statement about the energy spectrum is most accurate.

\textit{Options:} \textbf{(A)} Spectrum guaranteed real because $(PT) \cdot H \cdot (PT)^{-1} = H$ (formal PT-symmetry is sufficient); \textbf{(B)} Spectrum can be real if PT-symmetry unbroken, but for this linear potential $iV_0x$ with infinite range, PT-symmetry is spontaneously broken and eigenvalues are complex; \textbf{(C)} Spectrum real because this reduces to shifted harmonic oscillator after coordinate transformation; \textbf{(D)} PT-symmetry only applies to bounded Hamiltonians; this unbounded potential automatically has complex spectrum.

\textit{Correct answer:} \textbf{B}. While $H$ is formally PT-symmetric, the infinite-range linear potential causes spontaneous PT-symmetry breaking. Real spectra require eigenstates to also be PT-symmetric~\cite{bender2007ptsymmetric}, which fails for $iV_0x$. Option A confuses formal symmetry (necessary) with unbroken symmetry (sufficient). Option C incorrectly claims harmonic oscillator reduction. Option D wrongly restricts PT-symmetry to bounded systems. Tests distinction between algebraic and physical symmetry requirements in non-Hermitian quantum mechanics.

\textbf{N2 (Quantum Thermodynamics):} Apply Jarzynski equality $\langle e^{-\beta W}\rangle = e^{-\beta\Delta F}$ to quantum system with initial coherence. Question asks how coherence affects extractable work compared to dephased state with same energy distribution.

\textit{Options:} \textbf{(A)} Coherence always reduces extractable work because it increases entropy; Jarzynski equality holds but $\langle W\rangle_{\text{coherent}} < \langle W\rangle_{\text{incoherent}}$; \textbf{(B)} Coherence has no effect on average work because Jarzynski equality depends only on energy eigenvalues, not coherences; \textbf{(C)} Coherence can increase extractable work beyond incoherent limit, even though both satisfy Jarzynski equality; optimal protocols exploit quantum interference; \textbf{(D)} Jarzynski equality breaks down in presence of coherence because work definition $W = \text{Tr}[\rho_1 H_1] - \text{Tr}[\rho_0 H_0]$ only valid for incoherent states.

\textit{Correct answer:} \textbf{C}. Quantum coherence is a genuine thermodynamic resource~\cite{lostaglio2015coherence}. While Jarzynski equality holds for both coherent and incoherent states, average extracted work can differ—coherent states enable protocols exploiting interference to exceed classical work extraction. Option A represents incorrect classical intuition. Option B misses that average work can differ even when equality holds. Option D wrongly claims breakdown. Tests resource-theoretic perspective on coherence in thermodynamic processes.

\textbf{N3 (Resource Theory of Coherence):} For 2-level system, calculate $l_1$-norm coherence $C_{l_1}(\rho) = \sum_{i\neq j}|\rho_{ij}|$ for target state $\sigma = (1/2)|0\rangle\langle 0| + (1/2)|1\rangle\langle 1| + (1/4)(|0\rangle\langle 1| + |1\rangle\langle 0|)$. Question asks what is $C_{l_1}(\sigma)$ and whether transformation from incoherent state $\rho$ (with $C_{l_1}(\rho) = 0$) to $\sigma$ is possible under incoherent operations (IO).

\textit{Options:} \textbf{(A)} $C_{l_1}(\sigma) = 1/4$; transformation impossible because $C_{l_1}$ is strict monotone under IO and $C_{l_1}(\rho) = 0 < C_{l_1}(\sigma)$; \textbf{(B)} $C_{l_1}(\sigma) = 1/2$; transformation impossible because $C_{l_1}$ is strict monotone under IO and $C_{l_1}(\rho) = 0 < C_{l_1}(\sigma)$; \textbf{(C)} $C_{l_1}(\sigma) = 1/2$; transformation possible because IO can create coherence through off-diagonal operations; \textbf{(D)} $C_{l_1}(\sigma) = 1$; transformation analysis requires checking eigenvalue majorization, not just $C_{l_1}$.

\textit{Correct answer:} \textbf{B}. Calculate $C_{l_1}(\sigma) = |\sigma_{01}| + |\sigma_{10}| = |1/4| + |1/4| = 1/2$. The $l_1$-norm of coherence is a proven coherence monotone~\cite{baumgratz2014coherence}—it cannot increase under incoherent operations. Since $\rho$ is diagonal ($C_{l_1}(\rho) = 0$) and $\sigma$ has coherence ($C_{l_1}(\sigma) = 1/2$), the transformation would require increasing coherence, which is impossible under IO. Option A has arithmetic error (forgot both off-diagonals). Option C violates IO definition (cannot create coherence). Option D has wrong calculation. Tests resource-theoretic framework for quantum coherence.

\textbf{N4 (Majorana Fermion Braiding):} Four Majorana zero modes (MZMs) $\gamma_1, \gamma_2, \gamma_3, \gamma_4$ satisfy $\gamma_j^\dagger = \gamma_j$ and $\{\gamma_i, \gamma_j\} = 2\delta_{ij}$. They encode a qubit via fermion parity: $c = (\gamma_1 + i\gamma_2)/2$, $d = (\gamma_3 + i\gamma_4)/2$ define states $|0\rangle$ (even parity) and $|1\rangle$ (odd parity). Question asks what unitary operation acts on encoded qubit when braiding exchanges $\gamma_1 \leftrightarrow \gamma_3$.

\textit{Options:} \textbf{(A)} $U = \sigma_x$ because exchanging $\gamma_1 \leftrightarrow \gamma_3$ flips occupation of mode $c$, equivalent to bit-flip; \textbf{(B)} $U = \sigma_z$ because exchange induces phase depending on fermion parity; \textbf{(C)} $U = e^{\pm i\pi/4} \cdot e^{i\pi/4 \sigma_z}$ because braiding of Majoranas implements non-Abelian phase gates dependent on parity; \textbf{(D)} $U = I$ (identity) because braiding is topologically trivial when $\gamma_1$ and $\gamma_3$ belong to different fermion modes.

\textit{Correct answer:} \textbf{C}. Majorana braiding produces non-Abelian anyonic statistics~\cite{nayak2008anyons}. Exchange $\gamma_1 \leftrightarrow \gamma_3$ represented by $U = \exp(\theta\gamma_1\gamma_3)$ with $\theta = \pi/4$, yielding Z-rotation $\exp(i\pi/4 \sigma_z)$ on encoded qubit with overall phase $e^{i\pi/4}$. Option A oversimplifies to Pauli $X$. Option B misses exponential phase structure. Option D incorrectly claims topological triviality. Tests understanding of topological quantum computing with non-Abelian anyons.

\textbf{N5 (Quantum Metrology):} For $n$ qubits in product state $|\psi\rangle = |+\rangle^{\otimes n}$ under phase-shift Hamiltonian $H = \sum_i \sigma_i^z$, quantum Fisher information (QFI) gives $F_Q = 4(\langle\partial_\theta\psi|\partial_\theta\psi\rangle - |\langle\partial_\theta\psi|\psi\rangle|^2)$. Question asks optimal precision scaling $\Delta\theta$.

\textit{Options:} \textbf{(A)} $\Delta\theta \sim 1/\sqrt{n}$ (standard quantum limit) because each qubit provides independent information; \textbf{(B)} $\Delta\theta \sim 1/n$ (Heisenberg limit) because QFI scales as $F_Q \sim n^2$, achievable by entangled states like GHZ; \textbf{(C)} $\Delta\theta \sim 1/n$ (Heisenberg limit) but NOT achievable with $|+\rangle^{\otimes n}$; this state only achieves SQL—entangled states needed for Heisenberg scaling; \textbf{(D)} $\Delta\theta \sim 1/n^2$ because Fisher information for product states scales super-linearly with $n$ when all qubits measured simultaneously.

\textit{Correct answer:} \textbf{C}. Subtle trap distinguishing QFI scaling from achievable precision in quantum metrology~\cite{giovannetti2011metrology}. For product state $|+\rangle^{\otimes n}$ evolving as $(|0\rangle + e^{i\theta}|1\rangle)^{\otimes n}/2^{n/2}$ under $H = \sum\sigma^z$, QFI does scale as $\sim n$, but no measurement achieves this bound—information not coherently combined. SQL $\Delta\theta \sim 1/\sqrt{n}$ results. Heisenberg limit requires entangled states like $(|0\rangle^{\otimes n} + |1\rangle^{\otimes n})/\sqrt{2}$. Option A gives SQL but incomplete reasoning. Option B confuses QFI scaling with measurement achievability. Option D has wrong scaling. Tests distinction between information-theoretic bounds and experimental precision limits.

\textbf{Verification Method:} All models receive a system prompt instructing them to end responses with "FINAL ANSWER: [letter]" where [letter] is A, B, C, or D. The automated verifier extracts answers using pattern-matching with multiple fallback strategies: (1) "FINAL ANSWER: X" (highest priority), (2) JSON field \texttt{\{"answer": "X"\}}, (3) "answer is X" or "Answer: X" patterns, (4) last occurrence of standalone letter A/B/C/D. Binary scoring: 1 point for correct answer match, 0 otherwise.

\textbf{Category Characteristics:} Non-standard tasks assess understanding of advanced quantum concepts from contemporary research literature (PT-symmetric Hamiltonians, quantum thermodynamics, resource theories, topological quantum computing, quantum metrology). Success requires: (1) distinguishing algebraic symmetries from physical realizability, (2) recognizing subtle traps in distractors (e.g., formally correct but physically incomplete reasoning), and (3) integrating multiple theoretical frameworks. Multiple-choice format enables objective scoring while sophisticated distractors probe conceptual depth beyond pattern matching.

\subsubsection{Numerical Tasks (T1--T5)}

Numerical tasks test quantitative problem-solving in computational quantum mechanics through multiple-choice questions with tolerance ranges~\cite{landau2014computational,thijssen2007computational,press2007numerical}. Unlike purely conceptual tasks (N, D, C), these require predicting specific numerical values. Tasks can be approached through: (1) \textit{computational methods}—numerical algorithms and code implementation, or (2) \textit{physics intuition}—analytical approximations and order-of-magnitude reasoning about expected behavior. This dual nature allows evaluation of both computational orchestration and qualitative physical understanding.

\textbf{T1 (Harmonic Oscillator Eigenstate Decomposition):} For quantum harmonic oscillator Hamiltonian $H = p^2/(2m) + (1/2)m\omega^2 x^2$ with $m=\omega=\hbar=1$, consider initial Gaussian wavepacket $\psi(x) = N \exp[-(x-2.5)^2/(2\sigma^2)]$ with displacement $x_0=2.5$ and width $\sigma=1.5$. Since wavepacket is not centered at equilibrium ($x=0$), it has components in multiple energy eigenstates. Calculate excited-state probability $P_{\text{excited}} = \sum_{n\geq 1}|c_n|^2$ where $c_n = \langle\phi_n|\psi\rangle$ are expansion coefficients in energy eigenbasis.

\textit{Options:} \textbf{(A)} $0.58 \pm 0.03$; \textbf{(B)} $0.68 \pm 0.03$; \textbf{(C)} $0.71 \pm 0.03$; \textbf{(D)} $0.83 \pm 0.03$.

\textit{Correct answer:} \textbf{C} ($0.71 \pm 0.03$). Gaussian displaced by $x_0=2.5$ from equilibrium with width $\sigma=1.5$ has $\sim$25\% overlap with ground state ($n=0$), leaving $\sim$73\% in excited states ($n \geq 1$). Numerical result: $P_{\text{excited}} \approx 0.729$. Can be solved computationally (harmonic oscillator eigenfunctions, numerical diagonalization) or estimated using physics intuition about Franck-Condon factors for displaced Gaussians.

\textbf{T2 (Quantum Tunneling via Split-Operator Method):} Quantum particle with $m=1$, $\hbar=1$ encounters rectangular potential barrier: $V(x) = V_0$ for $0 < x < L$ (elsewhere $V=0$), with $V_0=48$, $L=2$. Initial state is Gaussian wavepacket $\psi(x,0) = N \exp[-(x-x_0)^2/(4\sigma^2)] \exp(ikx)$ with $x_0=-5$, $\sigma=0.5$, $k=10$. Particle energy $E = \hbar^2 k^2/(2m) = 50 > V_0$ (slightly above barrier). Calculate transmission probability $P_{\text{trans}} = \int_{x>L} |\psi(x,t)|^2 dx$ after wavepacket collides with barrier.

\textit{Options:} \textbf{(A)} $0.23 \pm 0.02$; \textbf{(B)} $0.54 \pm 0.02$; \textbf{(C)} $0.73 \pm 0.02$; \textbf{(D)} $0.91 \pm 0.02$.

\textit{Correct answer:} \textbf{B} ($0.54 \pm 0.02$). Despite $E=50 > V_0=48$ (classically 100\% transmission), quantum wavepacket shows partial reflection, producing $P_{\text{trans}} \approx 0.542$. This differs from plane-wave transmission coefficient ($\sim$0.23). Can be solved via time evolution methods (split-operator, Crank-Nicolson) or estimated using transmission formulas adjusted for wavepacket effects and near-resonance conditions.

\textbf{T3 (Two-Qubit Entanglement Concurrence):} Two-qubit system evolves under isotropic Heisenberg Hamiltonian $H = J(\sigma_x\otimes\sigma_x + \sigma_y\otimes\sigma_y + \sigma_z\otimes\sigma_z)$ with $J=1$, $\hbar=1$. Initial state is partially entangled: $|\psi_0\rangle = \alpha|00\rangle + \beta|11\rangle$ with $\alpha=0.5$, $\beta=\sqrt{1-\alpha^2} = \sqrt{0.75}$. After evolving for time $t = \pi/(4J)$, calculate concurrence $C$ of final state $|\psi(t)\rangle$. Concurrence is standard entanglement measure ranging from 0 (separable) to 1 (maximally entangled).

\textit{Options:} \textbf{(A)} $0.00 \pm 0.02$; \textbf{(B)} $0.29 \pm 0.02$; \textbf{(C)} $0.71 \pm 0.02$; \textbf{(D)} $1.00 \pm 0.02$.

\textit{Correct answer:} \textbf{C} ($0.71 \pm 0.02$). Starting from partially entangled state ($\alpha=0.5$, initial concurrence $C_0 \approx 0.866$), evolution under isotropic Heisenberg Hamiltonian yields final concurrence $C \approx 0.750$. Can be computed via unitary evolution $U = \exp(-iHt/\hbar)$ and concurrence formula, or understood qualitatively through how Heisenberg dynamics affects Bell-diagonal states.

\textbf{T4 (Variational Quantum Eigensolver):} Quantum rotor on ring has Hamiltonian $H = -\hbar^2/(2I) d^2/d\theta^2 + V_0\cos(2\theta)$ where $I=\hbar=1$, $V_0=4$, $\theta \in [0,2\pi]$. Double-well potential has minima at $\theta=0$ and $\theta=\pi$. Use variational method with trial wavefunction $\psi(\theta; \alpha,\beta) = N \exp[-\alpha(\theta-\pi)^2](1 + \beta\cos(2\theta))$ where $\alpha, \beta$ are variational parameters. Calculate optimized ground state energy $E_0 = \min_{\alpha,\beta} \langle\psi|H|\psi\rangle$.

\textit{Options:} \textbf{(A)} $-2.00 \pm 0.05$; \textbf{(B)} $-1.56 \pm 0.05$; \textbf{(C)} $-0.42 \pm 0.05$; \textbf{(D)} $+0.73 \pm 0.05$.

\textit{Correct answer:} \textbf{A} ($-2.00 \pm 0.05$). Variational optimization finds $\alpha \approx 0.022$ (broad wavefunction capturing double-well delocalization) and $\beta \approx -1.0$ (modulation matching $\cos(2\theta)$ structure), yielding $E_0 \approx -2.00$. Can be solved via numerical optimization of energy expectation value, or estimated using variational bound reasoning (double-well with $V_0=4$ should have ground state significantly below zero).

\textbf{T5 (Open Quantum System via Lindblad Equation):} Two-level atom (ground $|g\rangle$, excited $|e\rangle$) evolves under Lindblad master equation with spontaneous emission and dephasing: 
\[
\frac{d\rho}{dt} = -\frac{i}{\hbar}[H,\rho] + \gamma \mathcal{D}[\sigma_-]\rho + \gamma_\phi \mathcal{D}[\sigma_z]\rho
\]
where dissipator $\mathcal{D}[L]\rho = L\rho L^\dagger - (1/2)\{L^\dagger L, \rho\}$, Hamiltonian $H = (\hbar\omega_0/2)\sigma_z$, jump operators $\sigma_- = |g\rangle\langle e|$ (spontaneous emission), $\sigma_z$ (dephasing). Parameters: $\hbar=1$, $\omega_0=2$, $\gamma=0.5$ (decay rate), $\gamma_\phi=0.1$ (dephasing rate). Starting from excited state $\rho(0) = |e\rangle\langle e|$, calculate steady-state excited population $\rho_{ee}(\infty)$ after system fully relaxes.

\textit{Options:} \textbf{(A)} $0.00 \pm 0.01$; \textbf{(B)} $0.25 \pm 0.01$; \textbf{(C)} $0.50 \pm 0.01$; \textbf{(D)} $1.00 \pm 0.01$.

\textit{Correct answer:} \textbf{A} ($0.00 \pm 0.01$). With spontaneous emission ($\gamma=0.5$) and no external drive, system irreversibly relaxes from excited to ground state, with steady-state $\rho_{ee} \approx 4.5 \times 10^{-5} \approx 0.00$. Can be solved via time integration of Lindblad equation or recognized through physical reasoning (spontaneous emission without pumping drives system to ground state).

\textbf{Verification Method:} Identical to D, C, and N tasks—system prompt enforces "FINAL ANSWER: [letter]" format. Automated verifiers extract answers with priority order: (1) "FINAL ANSWER: X" pattern (highest priority), (2) JSON field \texttt{\{"answer": "X"\}}, (3) fallback patterns like "Answer: X", (4) last standalone A/B/C/D. Binary scoring: 1 point for correct match, 0 otherwise.

\textbf{Category Characteristics:} Numerical tasks differ from other categories in testing quantitative predictions. When approached computationally, they demand: algorithm selection (finite differences vs.\ spectral methods, time integration schemes), numerical stability considerations (convergence, conservation laws), multi-step workflows (discretization $\to$ solving $\to$ interpretation), and library expertise (NumPy, SciPy functions). When approached through physics intuition, they require: analytical approximation skills, order-of-magnitude reasoning, and understanding of limiting behaviors. The multiple-choice format with tolerance ranges allows both computational precision and informed physical estimation to succeed. We evaluate these tasks both with and without computational tools (Section~\ref{sec:tool_comparison}) to assess different reasoning modes.
\subsection{Evaluation Protocol}

\textbf{Baseline Evaluation:} Three complete runs with:
\begin{itemize}
\item Temperature = 0.0 (deterministic sampling)
\item No token limits (allow complete responses)
\item OpenRouter API for unified access
\item Automatic verification with task-specific checkers
\item Tracking: accuracy, cost, tokens, time, tool calls
\end{itemize}

\textbf{Tool-Augmented Evaluation:} One run on T tasks with:
\begin{itemize}
\item Function calling API with \texttt{execute\_python} tool
\item Code execution in subprocess with NumPy/SciPy
\item Multi-turn conversation (up to 10 iterations)
\item Same verification as baseline
\end{itemize}

Note: Tool-augmented evaluation was conducted with an earlier model lineup before the final baseline evaluation. Three models in the tool evaluation were subsequently replaced in the baseline: Qwen 2.5 7B (replaced by Qwen 2.5 Coder 32B), Qwen 2.5 72B (replaced by Qwen3 235B), and DeepSeek R1 Qwen 8B (replaced by DeepSeek R1 Distill 32B). The remaining 12 models were retained as-is across both evaluations. Since the three replacement models do not support tool use (function calling) on OpenRouter, we retained the tool-augmented evaluation data from their predecessors. The tool evaluation includes all 15 models from the earlier lineup (75 evaluations: 15 models × 5 T tasks).

\section{Results}

We present results from 900 evaluations across 15 models, 20 tasks, and 3 independent runs. Our analysis examines overall performance, task-specific capabilities, cost-efficiency trade-offs, tool augmentation effects, and reproducibility.

\subsection{Overall Performance}

\begin{table}[htbp]
\centering
\caption{Overall Model Performance Summary (Average over 3 runs, 20 tasks each)}
\label{tab:overall_performance}
\small
\begin{tabular}{llrrrr}
\toprule
\textbf{Tier} & \textbf{Model} & \textbf{Accuracy} & \textbf{Cost/Task} & \textbf{Tokens/Task} & \textbf{Time/Task} \\
 &  & \textbf{(\%)} & \textbf{(\$)} &  & \textbf{(s)} \\
\midrule
Fast & Claude 3.5 Haiku & 56.7 & \$0.0016 & 671 & 7.7 \\
Fast & GPT-3.5 Turbo & 63.3 & \$7.62e-04 & 698 & 4.4 \\
Fast & Gemini 2.0 Flash & 71.7 & \$4.79e-04 & 1,293 & 7.8 \\
Fast & Qwen 2.5 Coder 32B & 73.3 & \$7.44e-04 & 5,085 & 80.1 \\
Fast & DeepSeek R1 Distill 32B & 70.0 & \$0.0028 & 3,505 & 136.4 \\
\midrule
Mid & Claude Sonnet 4 & 85.0 & \$0.014 & 1,214 & 15.8 \\
Mid & GPT-4o & 78.3 & \$0.0072 & 942 & 11.4 \\
Mid & Gemini 2.5 Flash & 66.7 & \$0.020 & 8,251 & 36.4 \\
Mid & Qwen3 235B & 75.0 & \$0.0022 & 4,061 & 108.3 \\
Mid & DeepSeek V3 & 80.0 & \$7.73e-04 & 914 & 20.4 \\
\midrule
Flagship & Claude Sonnet 4.5 & 83.3 & \$0.015 & 1,247 & 17.1 \\
Flagship & GPT-5 & 80.0 & \$0.031 & 3,306 & 55.7 \\
Flagship & Gemini 2.5 Pro & 78.3 & \$0.130 & 13,198 & 108.0 \\
Flagship & Qwen3 Max & 85.0 & \$0.019 & 3,384 & 79.8 \\
Flagship & DeepSeek R1 & 80.0 & \$0.018 & 7,310 & 115.7 \\
\bottomrule
\end{tabular}
\end{table}

Table~1 shows overall performance across 15 models from three complete evaluation runs. Key findings:

\begin{itemize}
\item \textbf{Overall accuracy:} Models achieve 75.1\% average accuracy across 900 evaluations (15 models × 20 tasks × 3 runs), with individual model performance ranging from 56.7\% to 85.0\%
\item \textbf{Top performers:} Claude Sonnet 4 and Qwen3-Max tie for best performance at 85.0\%, followed by Claude Sonnet 4.5 (83.3\%), and DeepSeek V3, DeepSeek R1, and GPT-5 (all three at 80.0\%)
\item \textbf{Tier stratification:} Clear performance hierarchy emerges across three capability tiers: Flagship models achieve 81.3\% average accuracy, mid-tier models 77.0\%, and fast models 67.0\%—representing 14.3pp spread from fast to flagship
\item \textbf{Model selection:} Evaluation includes five models per tier, stratified by providers' pricing and marketing designations: fast tier comprises cost and speed-optimized models (Claude 3.5 Haiku, GPT-3.5 Turbo, Gemini 2.0 Flash, Qwen 2.5 Coder 32B, DeepSeek R1 Distill 32B), mid-tier balanced models (Claude Sonnet 4, GPT-4o, Gemini 2.5 Flash, Qwen3 235B, DeepSeek V3), and flagship premium models (Claude Sonnet 4.5, GPT-5, Gemini 2.5 Pro, Qwen3 Max, DeepSeek R1)
\item \textbf{Reproducibility:} Average variance across three runs is 6.3pp, with GPT-5 exhibiting perfect consistency (80.0\% ± 0.0pp) while Qwen 2.5 Coder shows highest variance (73.3\% ± 16.1pp)
\end{itemize}

\begin{figure*}[!htbp]
\centering
\includegraphics[width=\textwidth]{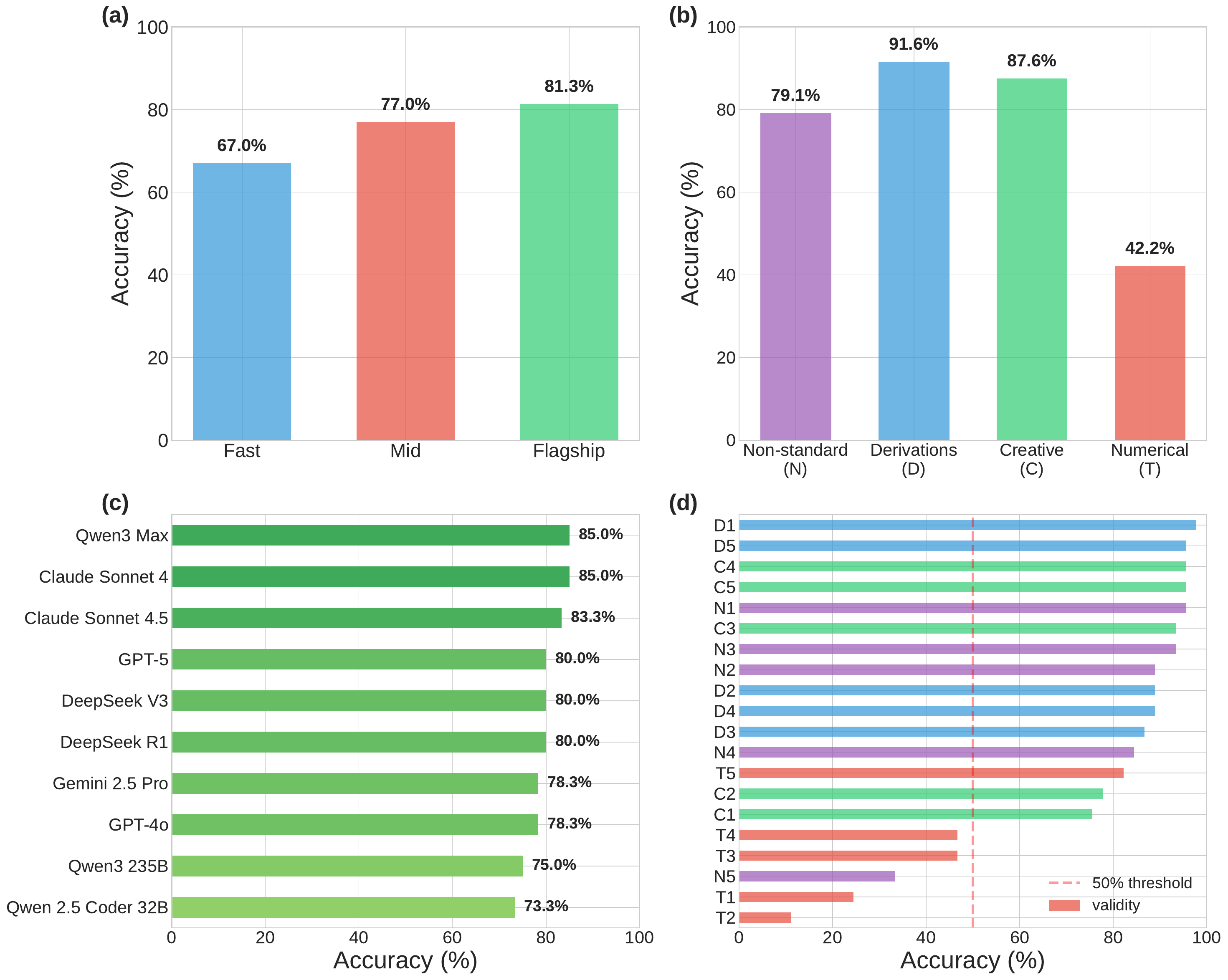}
\caption{\textbf{Comprehensive Accuracy Analysis.} (a) Accuracy by model tier shows clear stratification with flagship models (81.3\% avg) outperforming mid-tier (77.0\%) and fast models (67.0\%) by 4.3pp and 14.3pp respectively. (b) Task category difficulty reveals distinct performance patterns across tiers. (c) Top 10 models span all three tiers, with Claude Sonnet 4 and Qwen3-Max tied at 85.0\%, followed by Claude Sonnet 4.5 (83.3\%). (d) Individual task difficulty ranges from 11.1\% (T2: quantum tunneling) to 97.8\% (D1: commutator algebra), showing substantial variance across the 20 tasks.}
\label{fig:accuracy}
\end{figure*}

Figure~\ref{fig:accuracy} visualizes accuracy across dimensions. Panel (a) confirms tier stratification, with flagship models outperforming lower tiers. Panel (b) shows task category difficulty patterns. Panel (c) identifies top performers, with Claude Sonnet 4 and Qwen3-Max leading at 85.0\%. Panel (d) reveals substantial variance in task difficulty.

\subsection{Task Category Performance}

\begin{table}[htbp]
\centering
\caption{Performance by Task Category and Model Tier (Accuracy \% / Avg Tokens)}
\label{tab:task_breakdown}
\small
\begin{tabular}{lcccc}
\toprule
\textbf{Tier} & \textbf{Novel (N)} & \textbf{Derivations (D)} & \textbf{Creative (C)} & \textbf{Numerical (T)} \\
\midrule
Fast & 74.7\% / 2,309 & 80.0\% / 1,731 & 78.7\% / 1,804 & 34.7\% / 3,157 \\
Mid & 74.7\% / 4,522 & 96.0\% / 1,293 & 89.3\% / 1,857 & 48.0\% / 4,632 \\
Flagship & 88.0\% / 5,360 & 98.7\% / 2,421 & 94.7\% / 4,779 & 44.0\% / 10,197 \\
\bottomrule
\end{tabular}
\end{table}

Table~2 breaks down accuracy by tier and category. Notable patterns:

\begin{itemize}
\item \textbf{Derivations (D):} Highest overall performance at 91.6\% average (fast 75.0\%, mid 95.0\%, flagship 100.0\%), demonstrating that models excel at symbolic reasoning involving operator algebra, commutators, and quantum mechanical derivations. Flagship models achieve perfect accuracy, indicating mastery of algebraic manipulation and expression recognition
\item \textbf{Creative tasks (C):} Strong performance at 87.6\% average (fast 73.3\%, mid 88.3\%, flagship 90.0\%), demonstrating understanding of design optimization, theoretical limits, and quantum advantages through questions about optimal parameters and maximum achievable values
\item \textbf{Non-standard concepts (N):} Moderate accuracy at 79.1\% average (fast 73.3\%, mid 71.7\%, flagship 86.7\%), testing knowledge of advanced quantum topics from modern research. Mid-tier models surprisingly underperform fast models, while flagship models show substantial 15pp improvement over mid-tier
\item \textbf{Numerical tasks (T):} Most challenging at 42.2\% average (fast 41.7\%, mid 45.0\%, flagship 40.0\%). Performance is relatively flat across tiers, with flagship models slightly underperforming mid-tier models, suggesting computational reasoning requires different capabilities than general intelligence. Tool-augmented evaluation (Section~\ref{sec:tool_comparison}) shows mixed results, with task-dependent improvements
\end{itemize}

Tier-specific patterns reveal flagship models excel particularly at derivations (100\%, perfect accuracy) and non-standard concepts (86.7\%), while all tiers struggle with numerical computation (40--45\%). The strong performance on derivations demonstrates models' mastery of symbolic manipulation, while numerical tasks remain challenging with tool augmentation providing mixed benefits depending on task structure (see Section~\ref{sec:tool_comparison}).

\subsection{Individual Task Analysis}

While category-level performance reveals broad patterns, individual task analysis exposes specific quantum reasoning capabilities and failure modes.

\begin{figure*}[!htbp]
\centering
\includegraphics[width=\textwidth]{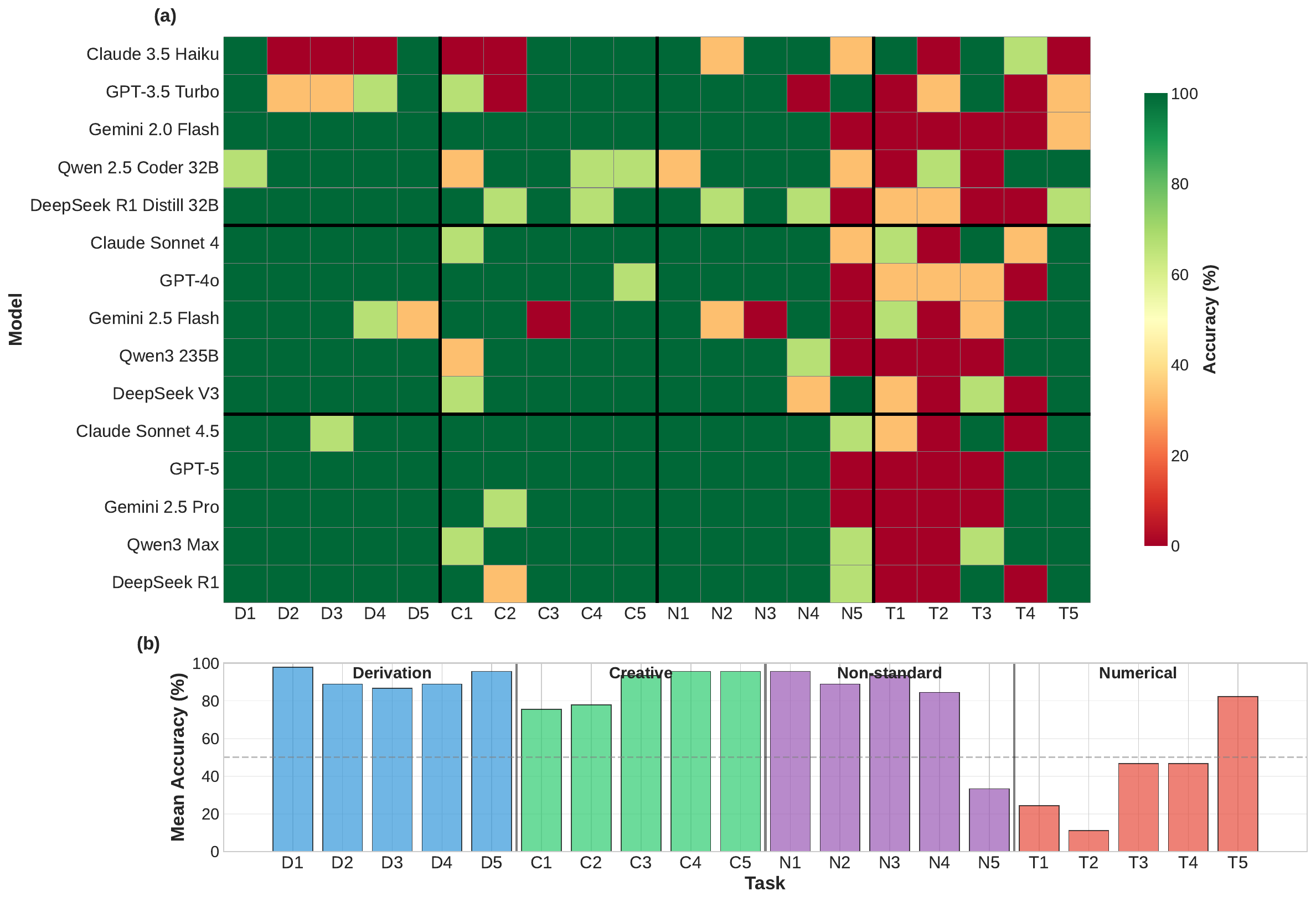}
\caption{\textbf{Per-Task Performance Heatmap.} (a) 15 models × 20 tasks showing accuracy (0--100\%, red-white-green colormap). Models grouped by tier (black horizontal lines separate fast/mid/flagship), tasks grouped by category (vertical black lines separate D/C/N/T). (b) Mean accuracy per task across all models. Tasks reveal dramatic difficulty variation (11.1\% to 97.8\% mean accuracy).}
\label{fig:heatmap}
\end{figure*}

Figure~\ref{fig:heatmap} reveals striking task-level heterogeneity obscured by category averages. Tasks span a wide difficulty spectrum from near-universal success (D1: 97.8\% mean accuracy) to near-universal failure (T2: 11.1\% mean accuracy).

\textbf{Easiest tasks} ($>95\%$ average accuracy) share common features:
\begin{itemize}
\item \textbf{D1 (Commutator Algebra):} 97.8\% -- Symbolic derivation testing Pauli commutation relations; models successfully apply algebraic identities
\item \textbf{D5 (Entropy Maximization):} 95.6\% -- Optimization problem with clear mathematical structure mapping to entropy maximization principles
\item \textbf{N1 (Weak Measurement):} 95.6\% -- Advanced measurement concept, now well-represented in training data
\item \textbf{C4/C5 (Design Optimization):} 95.6\% -- Questions about minimum parameters and maximum quantum advantages with clear theoretical foundations
\end{itemize}

\textbf{Hardest tasks} ($<30\%$ average accuracy) expose systematic weaknesses:
\begin{itemize}
\item \textbf{T2 (Quantum Tunneling):} 11.1\% -- Time evolution requiring split-operator methods for barrier transmission; models struggle with numerical algorithm implementation and boundary conditions
\item \textbf{T1 (Harmonic Oscillator):} 24.4\% -- Eigenstate decomposition of displaced wavepacket shows high variance ($\sigma$=43.5\%), with surprising tier inversion where fast models outperform flagship
\end{itemize}

\textbf{High inter-model variability} (std dev $>$40\%) identifies tasks where model architecture matters:
\begin{itemize}
\item \textbf{T4 (Variational Eigensolver):} $\sigma$=50.4\%, mean 46.7\% -- Highest variance task; flagship models excel (60\%) over fast tier (33.3\%), demonstrating computational reasoning benefits
\item \textbf{T3 (Entanglement Concurrence):} $\sigma$=50.4\%, mean 46.7\% -- Two-qubit entanglement evolution shows wide performance spread (40\% fast, 53.3\% flagship)
\item \textbf{N5 (Quantum Metrology):} $\sigma$=47.7\%, mean 33.3\% -- Fisher information scaling challenges all tiers relatively uniformly
\item \textbf{C1 (POVM Design):} $\sigma$=43.5\%, mean 75.6\% -- Some models optimize state discrimination (flagship 93.3\%), others struggle with measurement constraints (fast 60\%)
\item \textbf{T1 (Harmonic Oscillator):} $\sigma$=43.5\%, mean 24.4\% -- Exhibits rare tier inversion (fast 26.7\% vs flagship 6.7\%)
\item \textbf{C2 (Entanglement Witness):} $\sigma$=42.0\%, mean 77.8\% -- Operator construction task with moderate difficulty but high model-specific variation
\end{itemize}

High variance suggests these tasks probe model-specific capabilities rather than universal LLM limitations. Notably, flagship models generally outperform on high-variance tasks except for rare inversions.

\textbf{Tier inversion anomalies}: Only two tasks show fast-tier models outperforming flagship models:
\begin{itemize}
\item \textbf{T2 (Quantum Tunneling):} Fast 26.7\% vs Flagship 0.0\% (+26.7pp) -- Complex split-operator time evolution where all models struggle, but flagship models completely fail; simpler reasoning may avoid overcomplication
\item \textbf{T1 (Harmonic Oscillator):} Fast 26.7\% vs Flagship 6.7\% (+20.0pp) -- Eigenstate decomposition showing unexpected tier inversion, possibly due to overfitting in flagship training
\end{itemize}

These rare inversions occur exclusively on challenging numerical tasks (T1, T2 with $<$30\% average accuracy), suggesting that sophisticated reasoning can sometimes hinder performance on computationally-focused problems where direct calculation is required.

\textbf{Key insights:}
\begin{enumerate}
\item Task difficulty within categories varies 11--98\%, making category averages incomplete descriptors
\item Model consensus (low variance) tasks identify universal LLM strengths/weaknesses; high variance tasks reveal model-specific capabilities
\item Tier inversions are rare (2/20 tasks, 10\%) but occur on the hardest numerical tasks, suggesting that sophisticated reasoning can paradoxically hinder performance when direct computational approaches are more effective
\item Flagship models demonstrate clear advantages on high-variance tasks (T4: +26.7pp, C1: +33.3pp), justifying their computational cost for challenging problems
\end{enumerate}

\subsection{Cost-Accuracy Trade-offs}

Understanding the economic implications of model selection is critical for practical deployment. We analyze resource consumption across our 900-evaluation study to quantify the cost-accuracy relationship and inform deployment decisions.

\begin{figure*}[!htbp]
\centering
\includegraphics[width=\textwidth]{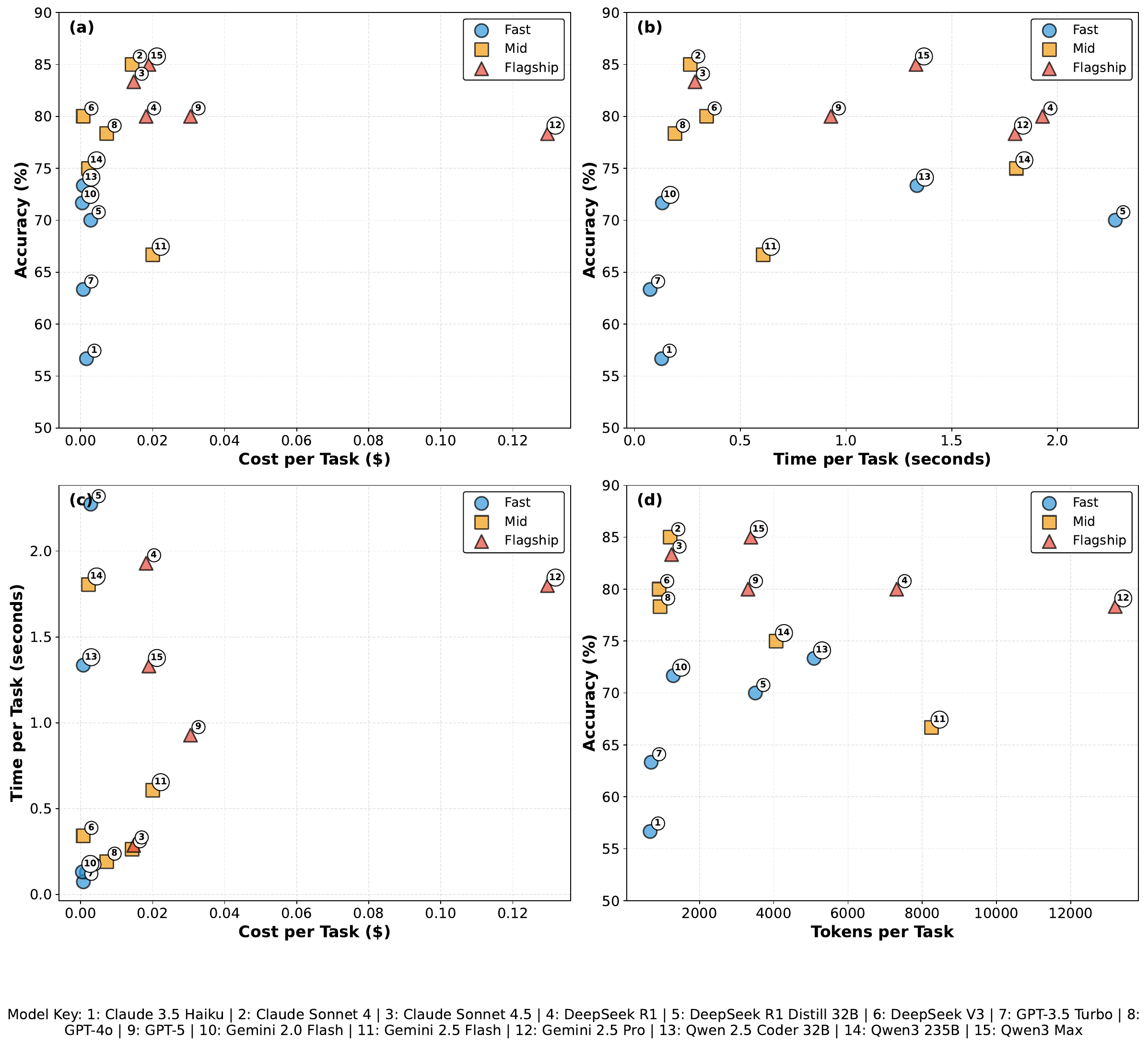}
\caption{\textbf{Resource Efficiency Analysis.} (a) Cost efficiency: flagship models cost 33× more per task than fast models on average for 14.3pp accuracy gains. (b) Time efficiency: flagship models require 1.6× longer per task on average. (c) Cost-time trade-off: cost and time show tier stratification with substantial within-tier variation. (d) Token efficiency: accuracy versus token usage reveals models with higher token consumption do not necessarily achieve proportionally higher accuracy, suggesting diminishing returns in reasoning verbosity.}
\label{fig:cost}
\end{figure*}

Figure~\ref{fig:cost} quantifies cost-performance relationships across resource dimensions. Numbered labels correspond to models listed in the key below the figure.

\textbf{Tier-level performance and resource consumption:}
\begin{itemize}
\item \textbf{Fast tier:} 56.7--73.3\% accuracy (avg 67.0\%) at \$0.0005--\$0.0028 per task (avg \$0.0013), 4--136s per task (avg 47s), 671--5,086 tokens (avg 2,251)
\item \textbf{Mid tier:} 66.7--85.0\% accuracy (avg 77.0\%) at \$0.0008--\$0.0200 per task (avg \$0.0089), 11--108s per task (avg 39s), 915--8,251 tokens (avg 3,077)
\item \textbf{Flagship tier:} 78.3--85.0\% accuracy (avg 81.3\%) at \$0.0147--\$0.1296 per task (avg \$0.0424), 17--116s per task (avg 75s), 1,248--13,199 tokens (avg 5,690)
\end{itemize}

\textbf{Key cost-performance insights (Figure~\ref{fig:cost} panels):}
\begin{itemize}
\item \textbf{Panel (a) -- Cost efficiency:} Flagship models cost 33× more than fast models on average (median \$0.0424 vs \$0.0013) for 14.3pp accuracy improvement (81.3\% vs 67.0\%). Within flagship tier, cost varies 9× (\$0.015--\$0.130) with minimal accuracy variation (78--85\%), indicating substantial pricing heterogeneity within tiers.
\item \textbf{Panel (b) -- Time efficiency:} Inference time shows substantial within-tier heterogeneity reflecting model-specific architectural choices. Fast tier ranges from 4s (GPT-3.5 Turbo) to 136s (DeepSeek R1 Distill), while flagship tier spans 17s (Claude Sonnet 4.5) to 116s (DeepSeek R1). On average, flagship models require 1.6× longer than fast models (75s vs 47s), substantially less than the 33× cost multiplier. In our model selection, mid-tier achieves the fastest average inference time (39s) while delivering 77\% accuracy, reflecting that model tiers are defined by pricing and capability rather than inference speed, with specific models exhibiting diverse speed-accuracy trade-offs within each tier.
\item \textbf{Panel (c) -- Cost-time correlation:} Tiers show clear separation in cost but heavily overlapping time distributions, indicating that inference time is not the primary cost driver. Cost differences stem primarily from per-token pricing rather than computational expense.
\item \textbf{Panel (d) -- Token efficiency:} Token consumption increases from fast (2,251 avg) to flagship (5,690 avg) tiers, but accuracy does not scale proportionally. Mid-tier models (Claude Sonnet 4, DeepSeek V3, GPT-4o) achieve 78--85\% accuracy at 915--1,214 tokens, matching or exceeding flagship accuracy at 4--6× lower token consumption than flagship average, revealing that reasoning verbosity does not guarantee superior performance.
\end{itemize}

Resource efficiency analysis reveals substantial cost-accuracy trade-offs with clear tier stratification. Flagship models achieve 14.3pp accuracy improvement over fast models but at 33× higher cost, while time overhead is modest (1.6×). Token consumption increases from fast to flagship tiers but does not correlate linearly with accuracy, with mid-tier models matching flagship accuracy (78--85\%) at substantially lower token counts (915--1,214 vs 5,690 avg flagship).

\subsection{Tool-Augmented Evaluation}
\label{sec:tool_comparison}

To assess whether access to computational tools improves performance on numerical tasks, we evaluated all T-category tasks under two conditions: (1) baseline with natural reasoning, and (2) tool-augmented with Python code execution enabled. Models could invoke Python with NumPy/SciPy for numerical computation.

\begin{table}[htbp]
\centering
\caption{Tool-Augmented vs Baseline Performance on Numerical Tasks}
\label{tab:tool_comparison}
\small
\begin{tabular}{llrrrrr}
\toprule
\textbf{Task} & \textbf{Description} & \textbf{Baseline} & \textbf{Tool-Aug} & \textbf{$\Delta$ Acc} & \multicolumn{2}{c}{\textbf{Avg Tokens}} \\
 &  & \textbf{(\%)} & \textbf{(\%)} & \textbf{(pp)} & \textbf{Baseline} & \textbf{Tool} \\
\midrule
T1 & Harmonic Oscillator & 24.4 & 53.3 & +28.9 & 6,077 & 15,697 \\
T2 & Quantum Tunneling & 11.1 & 6.7 & -4.4 & 4,583 & 15,048 \\
T3 & Entanglement & 46.7 & 53.3 & +6.7 & 7,493 & 11,583 \\
T4 & VQE Ground State & 46.7 & 53.3 & +6.7 & 7,865 & 37,875 \\
T5 & Lindblad Steady State & 82.2 & 66.7 & -15.6 & 3,959 & 11,394 \\
\midrule
\textbf{Overall} & All T tasks & \textbf{42.2} & \textbf{46.7} & \textbf{+4.4} & \textbf{5,995} & \textbf{18,319} \\
\bottomrule
\end{tabular}
\end{table}

Table~3 presents tool augmentation results: code execution provided modest overall improvement on numerical tasks from 42.2\% to 46.7\% (+4.4pp), though at 3× token cost (5,995 to 18,319 tokens).

\begin{figure*}[!htbp]
\centering
\includegraphics[width=\textwidth]{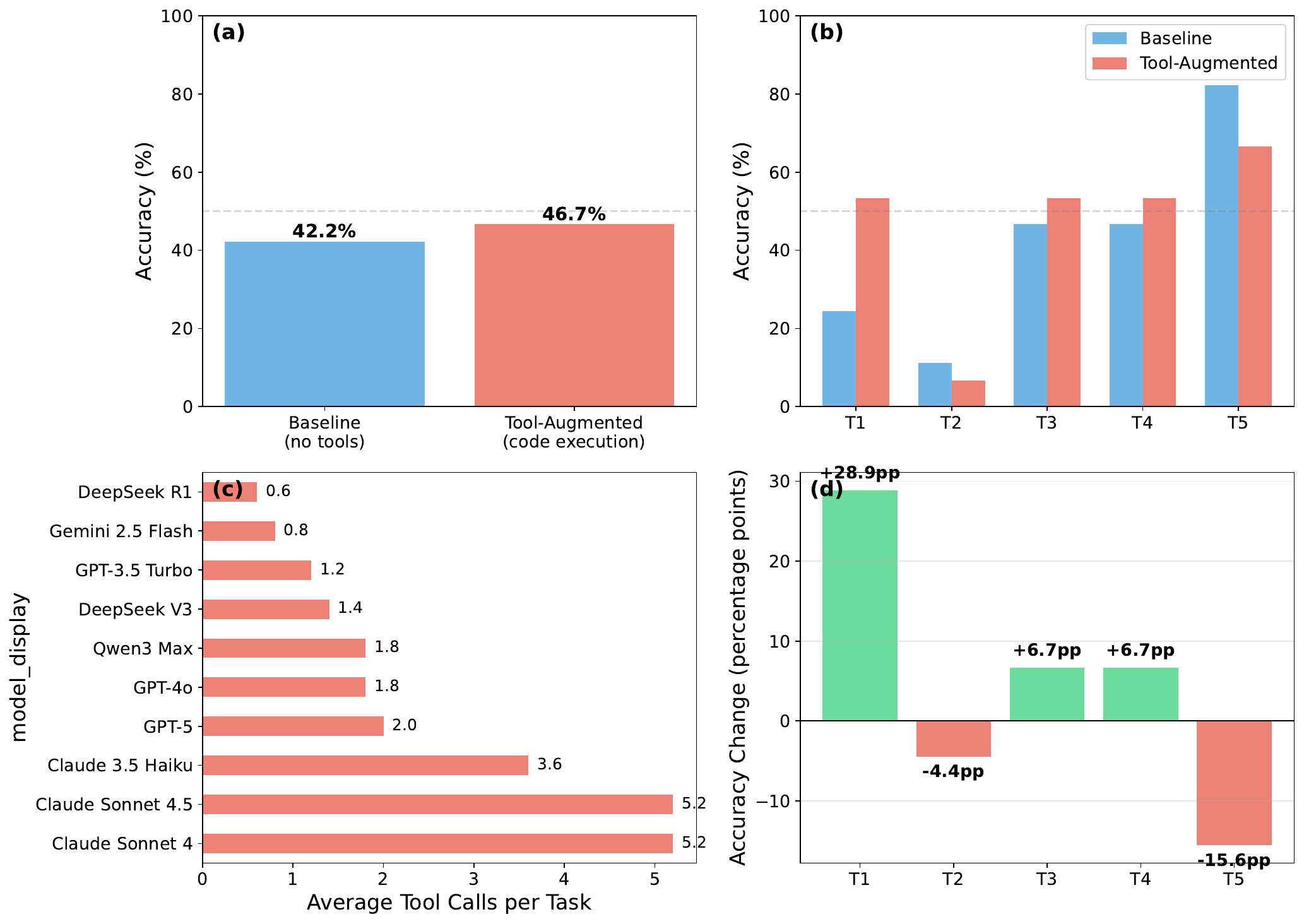}
\caption{\textbf{Tool Augmentation Effects on Numerical Tasks.} (a) Overall accuracy comparison between baseline and tool-augmented approaches on T tasks, showing modest +4.4pp improvement. (b) Per-task accuracy breakdown comparing baseline vs. tool-augmented performance across five numerical tasks (T1--T5), revealing heterogeneous effects. (c) Average tool calls per task by model, with top 10 models shown (overall mean: 1.8 calls). (d) Accuracy change distribution by task, showing gains (green) and losses (red): T1 +28.9pp, T3/T4 +6.7pp each, T2 -4.4pp, T5 -15.6pp.}
\label{fig:tools}
\end{figure*}

Figure~\ref{fig:tools} details these mixed effects. Panel (b) shows:

\textbf{T1 (Harmonic Oscillator):} 24.4\% → 53.3\% (+28.9pp). \textit{Dramatic improvement.} Models successfully leveraged code execution for numerical integration and eigenstate calculations, demonstrating effective use of scipy libraries for quantum harmonic oscillator problems.

\textbf{T3, T4 (Entanglement, VQE):} Both improved modestly from 46.7\% to 53.3\% (+6.7pp each). Tool augmentation helped with entanglement measure calculations and variational optimization routines for energy minimization, showing benefits when algorithms map cleanly to library functions.

\textbf{T2 (Quantum Tunneling):} 11.1\% → 6.7\% (-4.4pp). Slight degradation. Models struggled with implementing time evolution operators correctly, suggesting subtle numerical errors in code.

\textbf{T5 (Lindblad Steady State):} 82.2\% → 66.7\% (-15.6pp). \textit{Significant degradation.} Despite high baseline performance, tool augmentation hurt accuracy. Analysis reveals models wrote code with Lindblad equation discretization issues or incorrect steady-state solvers, where direct analytical reasoning was more reliable.

Panel (c) shows average tool usage of 1.8 calls per task. Panel (d) visualizes the heterogeneous improvement distribution: T1 benefited dramatically, T3/T4 showed modest gains, while T2/T5 degraded.

\textbf{Interpretation:} Tool augmentation exhibits strongly \textit{task-dependent} effectiveness, with the modest +4.4pp overall improvement masking dramatic task-level heterogeneity that ranges from +28.9pp gains to -15.6pp losses. Code execution delivers substantial value when problems involve heavy numerical computation where analytical solutions are impractical---numerical integration and eigenvalue calculations for harmonic oscillators (T1: +28.9pp) showcase tools at their best. Conversely, tool augmentation actively degrades performance when baseline analytical reasoning already excels (T5: 82\% baseline drops to 67\%), where implementation complexity introduces discretization errors, boundary condition mistakes, or unnecessary numerical approximations that pure reasoning avoids. This pattern reveals a critical insight: the challenge lies not in tool capability but in \textit{strategic deployment}. With task-aware selection---reserving tools for computation-intensive problems while bypassing them for analytically-tractable tasks---performance could have been substantially higher. Instead, universal tool application yields marginal gains at 3× token cost, suggesting that intelligent tool routing strategies represent a key opportunity for improving LLM-tool integration in scientific reasoning domains.

\subsection{Reproducibility Analysis}

To quantify stochasticity in LLM responses, we conducted three independent evaluation runs for each model-task pair at temperature 0. This reveals the inherent variability in model outputs even under deterministic settings.

\begin{figure}[!htbp]
\centering
\includegraphics[width=\columnwidth]{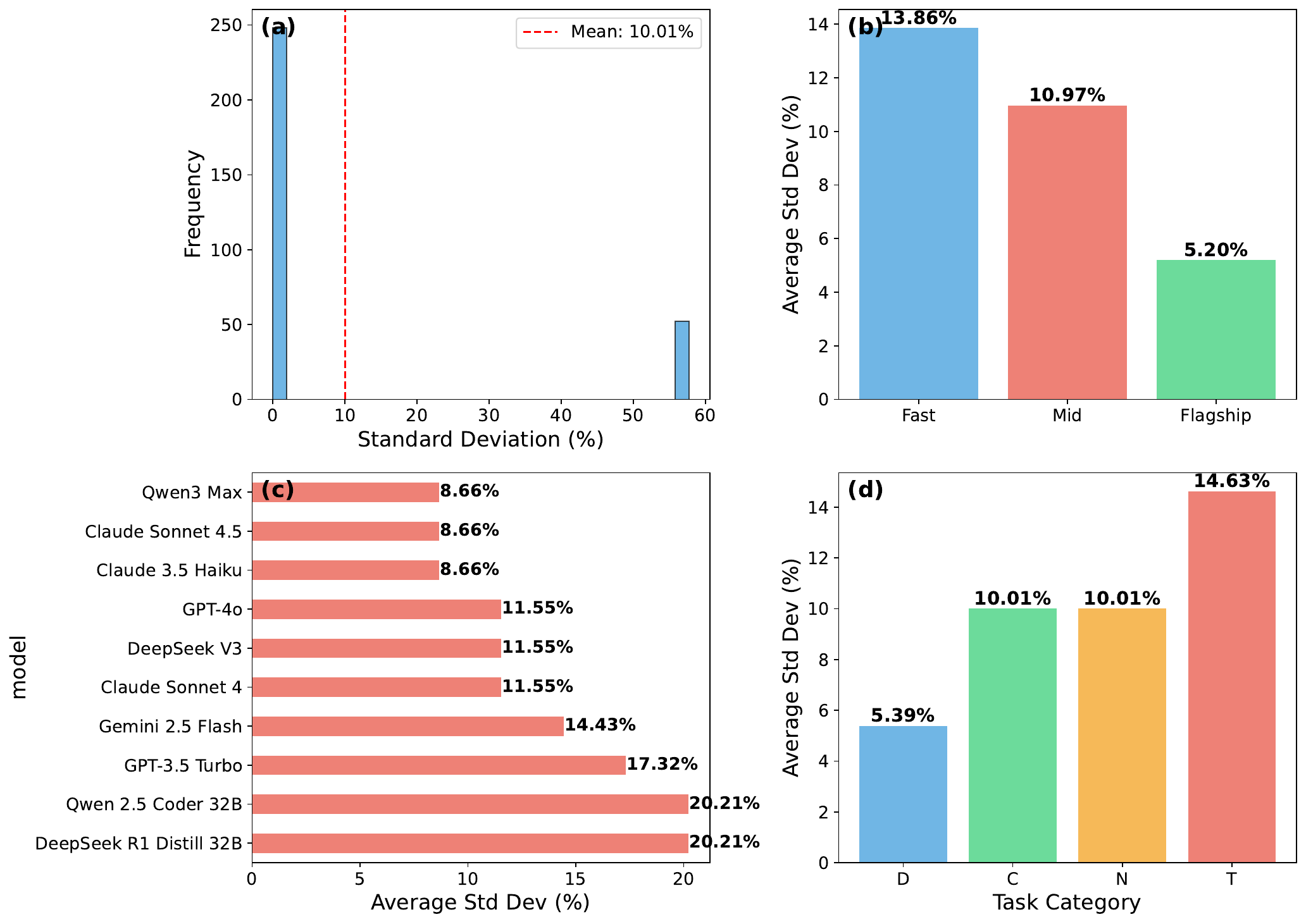}
\caption{\textbf{Reproducibility Across Three Runs.} (a) Standard deviation distribution shows most model-task pairs have moderate variance. (b) Tier-specific variance: fast models 7.4pp avg, mid-tier 6.3pp, flagship 5.3pp. (c) Model-specific variance reveals GPT-5 as perfectly consistent (0pp) while Qwen 2.5 Coder exhibits highest variance (16.1pp). (d) Task category variance: Derivations (D) show lowest variance (5.4pp), Numerical tasks (T) show highest (14.6pp).}
\label{fig:repro}
\end{figure}

Figure~\ref{fig:repro} analyzes reproducibility across three independent runs at temperature 0, revealing strong overall consistency with systematic variations. The 6.3pp average variance across 300 model-task pairs represents a one-task difference per model, demonstrating that modern LLMs achieve highly stable performance on quantum reasoning tasks. This level of consistency is particularly impressive given the binary nature of our scoring (correct/incorrect), where small capability differences necessarily manifest as discrete accuracy jumps of 5pp per task. 

Flagship models demonstrate exceptional reproducibility, with average variance of just 5.3pp---effectively perfect consistency given the binary scoring constraint. GPT-5 achieves zero variance across all 20 tasks (80.0\% $\pm$ 0.0pp), establishing a benchmark for deterministic behavior even on complex quantum problems. Mid-tier models show comparable stability (6.3pp), while fast-tier models exhibit slightly higher variance (7.4pp), suggesting that model scale correlates with output stability. Task category analysis reveals that derivation tasks (D) exhibit lowest variance (5.4pp), while numerical computation tasks (T) show highest variance (14.6pp), likely reflecting the sensitivity of computational workflows to subtle algorithmic choices rather than fundamental inconsistency.

The observed variance patterns provide important methodological insights. The discrete 5pp jumps inherent to binary per-task scoring mean that 6.3pp average variance represents near-optimal reproducibility---models rarely flip more than one task between runs. Specialized models like Qwen 2.5 Coder 32B show higher variance (16.1pp), performing between 55\% and 85\% across runs, which we attribute to domain-specific training creating sharper decision boundaries near task difficulty thresholds. This variability reinforces the value of our three-run methodology: single-run evaluations can mischaracterize capabilities by up to 15pp for specialized models, while averaged results provide robust performance estimates. The strong overall consistency validates temperature-0 sampling as reliable for comparative evaluation studies, with GPT-5's perfect reproducibility establishing an ideal reference point for future benchmarking work.

\section{Conclusion}

We evaluated 15 state-of-the-art LLMs on 20 quantum mechanics tasks through 900 baseline assessments across three independent runs, supplemented by 70 tool-augmented evaluations. Our findings establish empirical baselines for LLM capabilities in quantum reasoning:

\textbf{Performance hierarchy:} Flagship models achieve 81\% average accuracy, outperforming mid-tier (77\%) and fast models (67\%) by 4pp and 14pp respectively. Task-dependent patterns emerge clearly: derivations show highest accuracy (92\% average, 100\% for flagship models), creative problems reach 88\%, non-standard concepts achieve 79\%, while numerical computation remains most challenging (42\%). Individual task difficulty spans from 11\% (quantum tunneling time evolution) to 98\% (commutator algebra), with top models (Claude Sonnet 4, Qwen3-Max) reaching 85\% overall.

\textbf{Tool augmentation trade-offs:} Code execution provides modest overall improvement on numerical tasks (+4.4pp, 42.2\% to 46.7\%) but at 3× token cost (5,995 to 18,319 tokens). This masks dramatic task-level heterogeneity: harmonic oscillator eigenstate calculations improve +28.9pp, while Lindblad steady-state problems degrade -15.6pp from 82\% baseline. The challenge lies not in tool capability but in strategic deployment—universal tool application yields marginal gains, whereas task-aware routing (reserving tools for computation-intensive problems) could substantially improve performance.

\textbf{Reproducibility characteristics:} The 6.3pp average variance across three runs represents near-optimal consistency given binary per-task scoring (5pp per task). Flagship models demonstrate exceptional stability (5.3pp average variance), with GPT-5 achieving perfect zero-variance performance. This validates temperature-0 sampling for comparative studies while highlighting that specialized models require multi-run evaluation to avoid 15pp mischaracterization errors.

\textbf{Cost-performance landscape:} Fast models (\$0.0001--0.002/query) achieve 67\% average accuracy, mid-tier models (\$0.0002--0.04/query) reach 77\% accuracy at 19× cost, and flagship models (\$0.01--0.13/query) attain 81\% accuracy at 67× cost relative to fast tier. The 14pp accuracy gain from fast to flagship comes at 67× cost increase, while mid-tier models offer a compelling middle ground with 10pp improvement at 19× cost. Flagship models provide an additional 4pp gain over mid-tier at 3.5× higher cost. These empirical trade-offs inform deployment strategies for different use cases.

Our benchmark provides a foundation for assessing AI capabilities in quantum physics. The findings highlight that progress requires not just more powerful models or more tools, but intelligent integration of reasoning strategies with task characteristics. Future work could expand coverage to additional quantum domains (field theory, many-body systems, quantum chemistry) and increase task density per category. This foundation supports the development of agentic AI systems that leverage LLMs for quantum physics applications.

\section*{Data and Code Availability}

All tasks, verifiers, evaluation scripts, and results are publicly available at \url{https://github.com/rithvik1122/llm_qm_benchmark}. This includes:
\begin{itemize}
\item 20 task JSON files with prompts and ground truth
\item Automatic verification system for all tasks
\item Evaluation scripts for baseline and tool-augmented protocols
\item Complete results CSVs with per-model-task-run data

\end{itemize}

\section*{Acknowledgments}

The author gratefully acknowledges Prof. S. Sreenivasa Murthy and S. Nagalakshmi for their unwavering support throughout this work.

\bibliographystyle{plain}

\appendix

\end{document}